\documentclass{article}
\usepackage{kr}

\usepackage{times}
\usepackage{soul}
\usepackage{url}
\usepackage[hidelinks]{hyperref}
\usepackage[utf8]{inputenc}
\usepackage[small]{caption}
\usepackage{graphicx}
\usepackage{amsmath}
\usepackage{amsthm}
\usepackage{booktabs}
\usepackage{algorithm}
\usepackage{algorithmic}
\usepackage{listings}
\usepackage{amssymb}

\title{Benchmarking Defeasible Reasoning with Large Language Models - Initial Experiments and Future Directions}

\author{%
Ilias Tachmazidis$^1$\and
Sotiris Batsakis$^{2,1}$\and
Grigoris Antoniou$^3$\\
\affiliations
$^1$School of Computing and Engineering, University of Huddersfield, UK\\
$^2$Hellenic Mediterranean University, Greece\\
$^3$Leeds Beckett University, UK\\
\emails
i.tachmazidis@hud.ac.uk,
sbatsakis@hmu.gr,
g.antoniou@leedsbeckett.ac.uk
}

\begin{document}

\maketitle

\begin{abstract}
Large Language Models (LLMs) have gained prominence in the AI landscape due to their exceptional performance. Thus, it is essential to gain a better understanding of their capabilities and limitations, among others in terms of nonmonotonic reasoning. This paper proposes a benchmark that corresponds to various defeasible rule-based reasoning patterns. We modified an existing benchmark for defeasible logic reasoners by translating defeasible rules into text suitable for LLMs.
We conducted preliminary experiments on nonmonotonic rule-based reasoning using ChatGPT and compared it with reasoning patterns defined by defeasible logic.
\end{abstract}

\section{Introduction}
Large Language Models (LLMs) have caught people's attention recently due to the exceptional performance these systems  achieved in various language related tasks since they are the underlying technology behind chat bots such as ChatGPT\footnote{Available at: https://chat.openai.com/}. Large Language models such as LaMDA \cite{thoppilan2022lamda} and GPT \cite{openai2023gpt4} 
are based on training deep neural networks  with billions of parameters using  huge lexical datasets and often employing human judgment in a semi-supervised (e.g., reinforcement learning) training setting \cite{lambert2022illustrating,ouyang2022training}. The exceptional - often human level- performance of LLMs in various tasks has led to a widespread discussion about the potential benefits and dangers of such technologies in various areas and human society in general including petitions to pause research on more capable LLMs \cite{letters2023pause}. 
For example GPT-4 achieved human lever performance in various academic and professional exams   including a  score in the top 10\% of test takers  in the Uniform Bar Examination, this performance is attributed to a large degree to scaling  LLMs to larger training datasets and more complex models with larger number of parameters \cite{openai2023gpt4}.


Despite the impressive performance of Large Language Models, including their ability to demonstrate an emerging intelligent behaviour and  reasoning capabilities, leading to the point of considering them  forerunners of Artificial General Intelligence \cite{bubeck2023sparks} several issues related to LLMs have been identified, such as the energy cost of training LLMs \cite{luccioni2022estimating,strubell2019energy}, difficulty to control their behaviour \cite{luccioni2021s}, ensuring conformity with stakeholders requirements and norms, as well as interpreting their functionality \cite{bowman2023eight}. The interpetability of LLMs is a crucial issue since neural network based LLMs appear to be `black boxes', in contrast to logic based systems, and although various attempts exist to deal with this problem, including the use of LLMs to interpret LLMs \cite{bills2023language}, this is still an unresolved issue.    
In addition, since LLMs are trained on vast amounts of raw text they tend to replicate their input rather that apply robust reasoning \cite{bender2021dangers}. LLMs trained on raw text instead of structured knowledge bases integrating machine readable semantics contribute to the difficulty of achieving efficient reasoning and this is an issue examined in various works such as \cite{zhang2022paradox} and surveyed in \cite{huang2022towards}. Various attempts to  integrate  Knowledge Graphs (KGs) to LLMs have been proposed \cite{zhen2022survey,yin2022survey} as a solution to the last issue, but recent advances in LLMs capabilities, including high performance 
  on academic and professional exams \cite{openai2023gpt4}, 
   illustrated the need for an updated evaluation of the reasoning capabilities of LLMs. This updated evaluation should take into account the recent developments in the field, including the deployment of systems such as ChatGPT employing the benefits of scalability \cite{kaplan2020scaling} and the LLMs demonstrated ability to adjust to new tasks given just a small number of examples \cite{brown2020language}. Furthermore LLMs capabilities with respect to important formalisms such as defeasible reasoning have not been examined in detail yet. This kind of reasoning is important for cases where knowledge is incomplete and conflicting, which is the case in many application areas including law and healthcare. In previous work \cite{antoniou2023defeasible} preliminary experiments on LLM defeasible reasoning have been performed, but a systematic analysis involving benchmark construction containing several examples of different reasoning patters is missing.

This work is an initial step towards developing a deep understanding of reasoning capabilities of LLMs with emphasis of nonmonotonic reasoning. In order to achieve this we propose a benchmark for LLMs by modifying an existing benchmark for defeasible logic reasoners.
The proposed benchmark corresponds to various reasoning patterns that will be described in the following. Furthermore  we conducted preliminary experiments on nonmonotonic rule-based reasoning using ChatGPT and compared it with reasoning patterns defined by defeasible logic. 

\section{Background}
A defeasible theory D is a triple (F,R,$>$) where F is a finite set of facts (literals), R a finite set of rules,
and $>$ a superiority relation (acyclic relation upon R).

A rule r consists (a) of its antecedent (or body) A(r) which is a finite set of literals, (b) an arrow, and, (c) its
consequent (or head) C(r) which is a literal. There are three types of rules: strict rules, defeasible rules and
defeaters represented by a respective arrow $\rightarrow$, $\Rightarrow$ and $\leadsto$. Strict rules are rules in
the classical sense: whenever the premises are indisputable (e.g., facts) then so is the conclusion. Defeasible rules
are rules that can be defeated by contrary evidence. Defeaters are rules that cannot be used to draw any conclusions;
their only use is to prevent some conclusions.

Given a set R of rules, we denote the set of all strict rules in R by
R$_{s}$, and the set of strict and defeasible rules in R by R$_{sd}$. R[q] denotes the set of rules in R with consequent q.
If q is a literal, $\thicksim$q denotes the complementary literal (if q is a positive literal p then $\thicksim$q is $\neg$p;
and if q is $\neg$p, then $\thicksim$q is p).

A conclusion of D is a tagged literal and can have one of the following four forms:
\begin{itemize}
  \item $+\Delta$q, meaning that q is definitely provable in D.
  \item $-\Delta$q, meaning that we have proved that q is not definitely provable in D.
  \item $+\partial$q, meaning that q is defeasibly provable in D.
  \item $-\partial$q, meaning that we have proved that q is not defeasibly provable in D.
\end{itemize}

Provability is defined below. It is based on the concept of a derivation (or proof) in D = (F, R, $>$). A derivation is a
finite sequence P = P(1), ..., P(n) of tagged literals satisfying the conditions shown below. The conditions are essentially
inference rules phrased as conditions on proofs. P(1..$\imath$) denotes the initial part of the sequence P of length i.
For more details on provability and an explanation of the intuition behind the conditions below, see \cite{DBLP:journals/corr/cs-AI-0405090}.

\begin{tabbing}
$+\Delta$:	\=We may append P($\imath$ + 1) = $+\Delta$q if either \\
\> q $\in$ F or \\
\> $\exists$r $\in$ R$_{s}$[q] $\forall$$\alpha$ $\in$ A(r): $+\Delta$$\alpha$ $\in$ P(1..$\imath$)
\end{tabbing}

\begin{tabbing}
$-\Delta$: \=We may append P($\imath$ + 1) = $-\Delta$q if \\
\> q $\notin$ F and \\
\> $\forall$r $\in$ R$_{s}$[q] $\exists$$\alpha$ $\in$ A(r): $-\Delta$$\alpha$ $\in$ P(1..$\imath$)
\end{tabbing}

\begin{tabbing}
$+\partial$: \=We may append P ($\imath$ + 1) = $+\partial$q if either \\
\> (1) $+$\=$\Delta$q $\in$ P(1..$\imath$) or \\
\> (2)	\>(2.1) $\exists$r $\in$ R$_{sd}$[q] $\forall$$\alpha$ $\in$ A(r): $+\partial$$\alpha$ $\in$ P(1..$\imath$) and \\
\> \>(2.2) $-\Delta$ $\thicksim$q $\in$ P(1..$\imath$) and \\
\> \>(2.3) \=$\forall$s $\in$ R[$\thicksim$q] either \\
\> \>\>(2.3.1) $\exists$$\alpha$ $\in$ A(s): $-\partial$$\alpha$ $\in$ P(1..$\imath$) or \\
\> \>\>(2.3.\=2) $\exists$t $\in$ R$_{sd}$[q] such that \\
\> \>\>\>$\forall$$\alpha$ $\in$ A(t): $+\partial$$\alpha$ $\in$ P(1..$\imath$) and t $>$ s
\end{tabbing}

\begin{tabbing}
$-\partial$: \=We may append P($\imath$ + 1) = $-\partial$q if \\
\> (1) $-$\=$\Delta$q $\in$ P(1..$\imath$) and \\
\> (2)	\>(2.1) $\forall$r $\in$ R$_{sd}$[q] $\exists$$\alpha$ $\in$ A(r): $-\partial$$\alpha$ $\in$ P(1..$\imath$) or \\
\> \>(2.2) $+\Delta$ $\thicksim$q $\in$ P(1..$\imath$) or \\
\> \>(2.3) \=$\exists$s $\in$ R[$\thicksim$q] such that \\
\> \>\>(2.3.1) $\forall$$\alpha$ $\in$ A(s): $+\partial$$\alpha$ $\in$ P(1..$\imath$) and \\
\> \>\>(2.3.\=2) $\forall$t $\in$ R$_{sd}$[q] either \\
\> \>\>\>$\exists$$\alpha$ $\in$ A(t): $-\partial$$\alpha$ $\in$ P(1..$\imath$) or t $\ngtr$ s
\end{tabbing}

\section{Dataset}

We propose a dataset of scalable test theories which is inspired 
by~\cite{DBLP:journals/ijait/MaherRABM01}. In~\cite{DBLP:journals/ijait/MaherRABM01} authors focused on evaluating
the efficiency of existing defeasible reasoning systems. Here, we focus
on a translation of rules into text suitable for LLMs.
The proposed dataset is focused on typical defeasible inference patterns,
allowing a comparison between inputs for reasoning systems and LLMs.

\textbf{Empty.} First, we skip the \textbf{empty()} theory as it contains no facts, rules or
priorities. The \textbf{empty()} theory serves as a baseline for reasoning systems. However, there is no meaningful evaluation of a LLM
in the absence of text. 

\textbf{Chain.} Our first theory is \textbf{chain(n)}, where $a_{0}$ is at the end of a chain of \emph{n} rules $a_{i+1}$ $\Rightarrow$ $a_{i}$, with a single fact $a_{n}$ initiating the chain of inference (no priorities defined). For \textbf{chain(2)}, the defeasible rules are as follows:
\begin{lstlisting}
>> A0000002
r1: A0000002 => A0000001
r2: A0000001 => A0000000
\end{lstlisting}
Note that ``$>>$ A0000002'' denotes a fact following the syntax of SPINdle~\cite{rohaninezhad2015grounder}. Based on the fact \emph{A0000002} rule \emph{r1} infers that \emph{A0000001} is
deafeasibly provable, while rule \emph{r2} infers that \emph{A0000000} is
deafeasibly provable as well. In this work,
the structure of the theories is aimed at determining
through logical inference whether \emph{A0000000} is provable or not. Subsequently, the translation of \textbf{chain(2)} into plain text is as follows:
\begin{lstlisting}
A0000002 is an Arkon.
If A0000002 is an Arkon, then typically A0000001 is an Arkon.
If A0000001 is an Arkon, then typically A0000000 is an Arkon.
\end{lstlisting}
Notice the pattern, facts are expressed as
statements, while rules are expressed as 
if-then statements with the keyword ``typically'' denoting the defeasible nature
of the rule. Given the already identified affinity of ChatGPT to use other background
knowledge when predicates and atoms are real-world entities, we use imaginary names of
species on an imaginary planet, following~\cite{ford2000strategies}. Here, we
use ``Arkon'' in order to ask ChatGPT:
\begin{lstlisting}
Is A0000000 an Arkon?
\end{lstlisting}

The theory \textbf{chains(n)}, is a version of \textbf{chain(n)} with strict rules. For \textbf{chains(2)}, the defeasible rules are as follows:
\begin{lstlisting}
>> A0000002
r1: A0000002 -> A0000001
r2: A0000001 -> A0000000
\end{lstlisting}
The translation of \textbf{chains(2)} into plain text is as follows:
\begin{lstlisting}
A0000002 is an Arkon.
If A0000002 is an Arkon, then A0000001 is an Arkon.
If A0000001 is an Arkon, then A0000000 is an Arkon.
\end{lstlisting}
Notice the absence of keyword ``typically'' in the if-then statements. 

\textbf{Circle.} In defeasible logic, cyclical chains of reasoning
do not lead to inferences. More specifically, in the theory \textbf{circle(n)}, $a_{0}$ is part of a circle of \emph{n} rules $a_{i+1~mod~n}$ $\Rightarrow$ $a_{i}$ (no facts or priorities defined). For \textbf{circle(2)}, the defeasible rules are as follows:
\begin{lstlisting}
r1: A0000000 => A0000001
r2: A0000001 => A0000000
\end{lstlisting}
Due to the cyclical nature of the rules, no defeasible conclusion is infered for either \emph{A0000000} or \emph{A0000001}.
The translation of \textbf{circle(2)} into plain text is as follows:
\begin{lstlisting}
If A0000000 is an Arkon, then typically A0000001 is an Arkon.
If A0000001 is an Arkon, then typically A0000000 is an Arkon.
\end{lstlisting}

The theory \textbf{circles(n)}, is a version of \textbf{circle(n)} with strict rules. For \textbf{circles(2)}, the defeasible rules are as follows:
\begin{lstlisting}
r1: A0000000 -> A0000001
r2: A0000001 -> A0000000
\end{lstlisting}
The translation of \textbf{circles(2)} into plain text is as follows:
\begin{lstlisting}
If A0000000 is an Arkon, then A0000001 is an Arkon.
If A0000001 is an Arkon, then A0000000 is an Arkon.
\end{lstlisting}
Notice again the absence of keyword “typically” in the if-then statements.

\textbf{Directed Acyclic Graph (DAG).} In order to consider more complex inference structures, we define theory \textbf{dag(n,k)}, where $a_{0}$ is the root of a 
k-branching tree of depth \emph{nk} in which every literal occurs \emph{n} times. 
The inference process is initiated by \emph{k} facts, namely $a_{nk+1}$, ..., $a_{nk+k}$ (no priorities defined). For \textbf{dag(2,2)}, the defeasible rules are as follows:
\begin{lstlisting}
>> A0000006
>> A0000005
r1: A0000006, A0000005 => A0000004
r2: A0000005, A0000004 => A0000003
r3: A0000004, A0000003 => A0000002
r4: A0000003, A0000002 => A0000001
r5: A0000002, A0000001 => A0000000
\end{lstlisting}
Notice that \emph{nk + 1} (here 5) rules are generated, with \emph{k} (here 2) facts, namely \emph{A0000006} and \emph{A0000005} making rule \emph{r1} applicable, inferring \emph{A0000004}. By applying rules \emph{r1}, \emph{r2}, \emph{r3}, \emph{r4} and \emph{r5} we can eventually infer \emph{A0000000}.
The translation of \textbf{dag(2,2)} into plain text is as follows:
\begin{lstlisting}
A0000006 is an Arkon.
A0000005 is an Arkon.
If A0000006 is an Arkon and A0000005 is an Arkon, then typically A0000004 is an Arkon.
If A0000005 is an Arkon and A0000004 is an Arkon, then typically A0000003 is an Arkon.
If A0000004 is an Arkon and A0000003 is an Arkon, then typically A0000002 is an Arkon.
If A0000003 is an Arkon and A0000002 is an Arkon, then typically A0000001 is an Arkon.
If A0000002 is an Arkon and A0000001 is an Arkon, then typically A0000000 is an Arkon.
\end{lstlisting}
Notice that multiple predicates in the body of a rule are connected through the keyword ``and'' in the if part of the if-then statement.

\textbf{Levels.} All mentioned theories above contained no conflicts. However, conflict
resolution is an integral part of defeasible reasoning. Thus, theory \textbf{levels-(n)}
 defines a cascade of \emph{n} conclusions, namely there are rules $true$ $\Rightarrow$ $a_{i}$
and $a_{i+1}$ $\Rightarrow$ $\neg$$a_{i}$, for $0 \leq i < n$ (no facts or priorities defined). For \textbf{levels-(2)}, the defeasible rules are as follows:
\begin{lstlisting}
r1: => A0000001
r2: A0000002 => -A0000001
r3: => A0000000
r4: A0000001 => -A0000000
\end{lstlisting}
Notice that when the body of a rule (here \emph{r1} and \emph{r3}) 
is empty then syntactically all preconditions are
considered to be met. Negative conclusions such as $\neg$\emph{A0000001} are prefixed with the minus sign, namely \emph{-A0000001}.
Since there is no fact supporting \emph{A0000002}, rule \emph{r2} does not apply, thus we conclude
\emph{A0000001} based on rule \emph{r1}. Subsequently, since both rules \emph{r3} and \emph{r4}
apply, we cannot conclude \emph{A0000000}. Notice an emerging pattern where \emph{A0000000} cannot be proved for even \emph{n}, while \emph{A0000000} can be proved for odd \emph{n} (due to alternating conflicts on subsequent levels).
The translation of \textbf{levels-(2)} into plain text is as follows:
\begin{lstlisting}
A0000001 is typically an Arkon.
If A0000002 is an Arkon, then typically A0000001 is not an Arkon.
A0000000 is typically an Arkon.
If A0000001 is an Arkon, then typically A0000000 is not an Arkon.
\end{lstlisting}

The theory \textbf{levels(n)}, is a version of \textbf{levels-(n)}, where
in addition there are superiority statements stating that, for odd \emph{i}, 
rule $a_{i+1}$ $\Rightarrow$ $\neg$$a_{i}$ is superior to $true$ $\Rightarrow$ $a_{i}$ (introducing $n/2$ priorities). 
For \textbf{levels(2)}, the defeasible rules are as follows:
\begin{lstlisting}
r1: => A0000001
r2: A0000002 => -A0000001
r2 > r1
r3: => A0000000
r4: A0000001 => -A0000000
\end{lstlisting}
Notice that due to the priority rule, if rule \emph{r2}
was applicable (e.g. with \emph{A0000002} given as fact) then $\neg$\emph{A0000001} would have been inferred (instead of 
\emph{A0000001} inferred here).
The translation of \textbf{levels(2)} into plain text is as follows:
\begin{lstlisting}
A0000001 is typically an Arkon, unless A0000002 is also an Arkon (namely then A0000001 is not an Arkon).
A0000000 is typically an Arkon.
If A0000001 is an Arkon, then typically A0000000 is not an Arkon.
\end{lstlisting}
A similar inference pattern emerges where \emph{A0000000} cannot be proved for even \emph{n}, while \emph{A0000000} can be proved for odd \emph{n} (even though the process of conflict resolution on subsequent levels is different for \textbf{levels(n)} compared to \textbf{levels-(n)}).

\textbf{Hierarchies.} The authors of~\cite{DBLP:journals/ijait/MaherRABM01} defined theories: (i)
\textbf{tree(n,k)}, where $a_{0}$ is the root of a k-branching tree of depth \emph{n} in which every literal occurs once, and (ii) \textbf{teams(n)}, where every literal is disputed, with two rules for 
$a_{i}$ and two rules for $\neg$$a_{i}$, and the rules for 
$a_{i}$ are superior to the rules for $\neg$$a_{i}$ (this situation is repeated recursively to a 
depth \emph{n}). In this work, we define \textbf{hierarchies(n,k)}, where $a_{0}$ is the root of a k-branching tree of depth \emph{n} in which every literal occurs once. In addition, every literal (internal node of the tree) is disputed, with $k/2$ rules for 
$a_{i}$ and $k/2$ rules for $\neg$$a_{i}$, where \emph{k} is even, and the rules for $a_{i}$ are superior to the rules for $\neg$$a_{i}$. Each external node of the tree
is a fact, namely there are $k^{n}$ facts.
For \textbf{hierarchies(2,2)}, the defeasible rules are as follows:
\begin{lstlisting}
>> A0000006
>> A0000005
>> A0000004
>> A0000003
r1: A0000006 => A0000002
r2: A0000005 => -A0000002
r1 > r2
r3: A0000004 => A0000001
r4: A0000003 => -A0000001
r3 > r4
r5: A0000002 => A0000000
r6: A0000001 => -A0000000
r5 > r6
\end{lstlisting}
The translation of \textbf{hierarchies(2,2)} into plain text is as follows:
\begin{lstlisting}
A0000006 is an Arkon.
A0000005 is an Arkon.
A0000004 is an Arkon.
A0000003 is an Arkon.
If A0000006 is an Arkon, then typically A0000002 is an Arkon.
If A0000005 is an Arkon, then typically A0000002 is not an Arkon, unless A0000006 is also an Arkon (namely then A0000002 is an Arkon).
If A0000004 is an Arkon, then typically A0000001 is an Arkon.
If A0000003 is an Arkon, then typically A0000001 is not an Arkon, unless A0000004 is also an Arkon (namely then A0000001 is an Arkon).
If A0000002 is an Arkon, then typically A0000000 is an Arkon.
If A0000001 is an Arkon, then typically A0000000 is not an Arkon, unless A0000002 is also an Arkon (namely then A0000000 is an Arkon).
\end{lstlisting}
Here, \emph{A0000000} can be proved for any given parameter \emph{n} and \emph{k} since
conflicts are always resolved in favour of $a_{i}$.

We consider out of scope of this work theory \textbf{mix(m,n,k)} from~\cite{DBLP:journals/ijait/MaherRABM01}, where there are \emph{m} defeasible rules for   
$a_{0}$ and \emph{m} defeaters against $a_{0}$, where each rule has \emph{n} atoms in its body (each atom
can be established by a chain of strict rules of length \emph{k}).

Table~\ref{test_theories_sizes} summarises the number of facts, rules and priorities generated
for each theory as a function of given parameters. Notice that the numbers in Table~\ref{test_theories_sizes}
do not necessarily match the numbers in~\cite{DBLP:journals/ijait/MaherRABM01} as we have
modified theory definitions.

\begin{table}[!ht]
    \centering
    \caption{Sizes of scalable theories.}
    \resizebox{\columnwidth}{!}{%
    \begin{tabular}{llll} 
    \toprule
    \textbf{Theory} & \textbf{Facts} & \textbf{Rules} & \textbf{Priorities} \\
    \midrule
    \textbf{chain(n)} & $1$ & $n$ & $0$ \\
    \textbf{chains(n)} & $1$ & $n$ & $0$ \\
    \textbf{circle(n)} & $0$ & $n$ & $0$ \\
    \textbf{circles(n)} & $0$ & $n$ & $0$ \\
    \textbf{dag(n,k)} & $k$ & $nk+1$ & $0$ \\
    \textbf{levels-(n)} & $0$ & $2n$ & $0$ \\
    \textbf{levels(n)} & $0$ & $2n$ & $n/2$\\
    \textbf{hierarchies(n,k)}  & $k^{n}$ & $k\sum_{i=0}^{n-1} k^{i}$  & $\frac{k}{2} \sum_{i=0}^{n-1} k^{i}$\\
    \bottomrule
    \end{tabular}
    }
    \label{test_theories_sizes}
\end{table}

\section{Experimental Results}

The proposed dataset is scalable, namely increasingly larger theories 
can be generated for increasing values of parameters \emph{n} and \emph{k}.
However, as a first step in this work, we focus on relatively small and
readable theories in order to assess empirically the inference patterns
of ChatGPT. We used GPT-4o in order to assess theories:
\textbf{chain(8)}\footnote{https://chatgpt.com/share/d8819744-9d90-42cb-aeba-95a28769f08e},
\textbf{chains(8)}\footnote{https://chatgpt.com/share/15f3f0a5-e8c9-4156-9050-7a28b62cc189},
\textbf{circle(8)}\footnote{https://chatgpt.com/share/a73155a9-9c3d-498d-8b2d-9532a9cd6d54},
\textbf{circles(8)}\footnote{https://chatgpt.com/share/3b381c57-f881-4631-92e0-3974c446a5df},
\textbf{dag(3,2)}\footnote{https://chatgpt.com/share/db79edc0-1f1e-4fec-bba1-b16fb6bd15a8},
\textbf{levels-(5)}\footnote{https://chatgpt.com/share/f332e764-91df-4672-b6d0-1fa95196c96e},
\textbf{levels(5)}\footnote{https://chatgpt.com/share/a1b2ae89-8f49-4f25-a7ea-fc94ac3e942d},
\textbf{hierarchies(2,4)}\footnote{https://chatgpt.com/share/bc1181d4-1f94-4a2f-8c9b-cf11fb28c35d}. ChatGPT was given the following
instructions for each theory:
\begin{lstlisting}
You are an expert on defeasible reasoning. Your task is to make logical conclusions based on provided knowledge (delimited with XML tags).
\end{lstlisting}
In addition, each prompt was based on the following template 
(namely, ``\{theory\}'' was substituted with each evaluated theory):
\begin{lstlisting}
Based on the following knowledge alone:

<knowledge>
``{theory}''
</knowledge>

Is A0000000 an Arkon?

Let's think step by step.
\end{lstlisting}

Each generated theory, such as \textbf{chain(8)}, was evaluated over 
four settings: 
\begin{itemize}
  \item \emph{A0000000} is not an Arkon with statements provided in random order (\emph{$-\partial$-rand}),
  \item \emph{A0000000} is an Arkon with statements provided in random order (\emph{$+\partial$-rand}),
  \item \emph{A0000000} is not an Arkon with statements provided in sequential order (\emph{$-\partial$-seq}),
  \item \emph{A0000000} is an Arkon with statements provided in sequential order (\emph{$+\partial$-seq}).
\end{itemize}

We evaluated statements provided in random order first in order to observe
any differences in generated responces when the same theory is provided in 
sequential order. Notice that ChatGPT conversations provided as links in footnotes
contain the four settings in the following order: \emph{$-\partial$-rand}, 
\emph{$+\partial$-rand}, \emph{$-\partial$-seq}, \emph{$+\partial$-seq}.
Results are summarised in Table~\ref{chatgpt_results}. 

\begin{table}[!ht]
    \centering
    \caption{ChatGPT inference results (for results annotated with $\dagger$ readers are referred to comments included in the main text).}
    \resizebox{\columnwidth}{!}{%
    \begin{tabular}{lllll} 
    \toprule
    \textbf{Theory} & \emph{$-\partial$-rand} & \emph{$+\partial$-rand} & \emph{$-\partial$-seq} & \emph{$+\partial$-seq} \\
    \midrule
    \textbf{chain(8)} & Correct & Correct & Correct & Correct \\
    \textbf{chains(8)} & Correct & Correct & Correct & Correct \\
    \textbf{circle(8)} & Correct & Correct & Correct & Correct \\
    \textbf{circles(8)} & Correct$\dagger$ & Correct & Correct & Correct \\
    \textbf{dag(3,2)} & Error$\dagger$ & Correct$\dagger$ & Correct & Correct \\
    \textbf{levels-(5)} & Error$\dagger$ & Error$\dagger$ & Error$\dagger$ &  Error$\dagger$ \\
    \textbf{levels(5)} & Error$\dagger$ & Correct$\dagger$ & Error$\dagger$ & Correct$\dagger$ \\
    \textbf{hierarchies(2,4)} & Correct & Correct & Correct & Correct \\
    \bottomrule
    \end{tabular}
    }
    \label{chatgpt_results}
\end{table}

For theory \textbf{chain(8)} we notice that \emph{$+\partial$-rand} and 
\emph{$+\partial$-seq} have similar inference patterns, namely
starting from provided facts, each rule is applied until a final
conclusion is reached (interestingly, statements provided in random
order do not change the inference sequence). We also notice  
that while \emph{$-\partial$-seq} evaluates
all rules sequentially (even after encountering \emph{A1111113}),
the inference of \emph{$-\partial$-rand} moves backwards once
\emph{A1111113} is encountered. Theory \textbf{chains(8)} follows
similar inference patterns based on if-then statements. 

For theory \textbf{circle(8)} we notice that for both 
\emph{$-\partial$-rand} and \emph{$-\partial$-seq} the circle
is identified and no conclusion can be drawn. For 
\emph{$+\partial$-rand} and \emph{$+\partial$-seq} we introduced
a fact that proves the circle, inference started from the given
fact leading to the inference of \emph{A0000000}.
Theory \textbf{circles(8)} follows similar inference patterns based on 
if-then statements. However, for \emph{$-\partial$-rand} the circle was
not explicitly identified as justification for inference.

For theory \textbf{dag(3,2)}, \emph{$-\partial$-rand} exhibited unusual
patterns (i.e. a potential hallucination), namely \emph{A1111114} was replaced with \emph{A0000004} 
(incorrectly), leading to the inference of \emph{A0000000} (while a rule based on \emph{A0000005} and \emph{A0000003} leading to \emph{A0000002} was not given as input).
Interestingly, for \emph{$-\partial$-seq} the inference pattern was
correct with \emph{A1111114} breaking the chain of inference (this
indicates that the sequence of statements can have an effect on
the inference process, which is not the case for standard reasoners).
The inference pattern for \emph{$+\partial$-rand} was correct, even
though it is unclear why the ``Chain of Reasoning'' did not include
the conjunction of two premises.  The inference pattern for 
\emph{$+\partial$-seq} was correct, with a well formed ``Chain of
 Reasoning''.

Theory \textbf{levels-(5)} introduces conflicting rules, with explanations
provided not matching expected inference for defeasible reasoning. This
might be attributed to the fact that ChatGPT starts from \emph{A0000000}
working backwards, while the lack of priorities over conflicting rules
introduces confusion. It is worth pointing out that \emph{$-\partial$-seq} 
introduced the statement ``A0000002 is typically an Arkon.'' (a fact not given as input), i.e. a potential hallucination.

Theory \textbf{levels(5)} contains priority rules, which provide
some clarity. However, for \emph{$-\partial$-rand}, since there are 
no priority rules for \emph{A0000002} the inference does not match deafeasible reasoning (it seems that ChatGPT does not follow the 
notion of defeasible reasoning where both \emph{A0000002} and $\neg$\emph{A0000002} might not be provable). 
Conversely, for \emph{$+\partial$-rand} the
combination of priorities and rules with failing premises (namely,
the resolution of \emph{A0000002} knowing that \emph{A0000003} is
not an Arkon) leads to conclusions that follow defeasible reasoning.
Interestingly, the inference steps of \emph{$-\partial$-seq} are
well structured, closely resembling defeasible reasoning (with the
exception of resolving \emph{A0000002} without a clear priority,
which leads to an error). The inference of \emph{$+\partial$-seq}
showed that the sequence of statements (where well structured
sequences lead to more intuitive inference steps) as well as the
presence of priorities and rules with failing premises can affect the inference process.

For theory \textbf{hierarchies(2,4)}, \emph{$-\partial$-rand} and
\emph{$+\partial$-rand} lead to correct inferences due to the presence 
of priorities and rules with failing premises. However, due to the
random order of statements, the explanation of inferences can
be challenging to follow. This is not the case for \emph{$-\partial$-seq} 
and \emph{$+\partial$-seq}, which exhibited correct and well
structured inference steps.

Overall, the following observations can be made:
\begin{itemize}
  \item ChatGPT seems to adopt the closed-world assumption, where facts are considered as true and missing information as false,
  \item Monotonic rules are applied leading to new conclusions,
  \item Conflicting rules are resolved in the presence of priorities and rules with failing premises,
  \item The presence of conflicting rules supporting both \emph{p} and \emph{$\neg$p}, with no clear priority, does not lead to a conclusion that neither \emph{p} nor \emph{$\neg$p} can be inferred,
  \item Additional facts or rules (not provided as input) could be automatically introduced (i.e. a potential hallucination),
  \item A well structured sequence of statements (given as input) 
  increases the readability of the inference process (compared to equivalent theories structured as statements in random order).
\end{itemize}

\section{Conclusion}
This work is a first step towards gaining a better understanding of reasoning capabilities of LLMs with respect to nonmonotonic reasoning. We proposed a benchmark tailored to LLMs through the modification of an existing benchmark for defeasible logic reasoners.
A range of reasoning patterns was covered by the proposed benchmark. Preliminary experiments 
indicated encouraging results for monotonic reasoning as well as certain challenges in 
the context of nonmonotonic rule-based reasoning. Future work will focus on expanding our
exploration of reasoning patterns that might pose a challenge to LLMs. Furthermore, while
this work was focused on small and readable theories, future efforts will examine the
effect of increasingly larger theories on the reasoning process of LLMs.

\bibliographystyle{kr}
\bibliography{main}

\end{document}